\documentclass[sigconf]{acmart}

\usepackage{todonotes}
\usepackage{lipsum}
\usepackage{multirow}
\usepackage{bm}
\usepackage{relsize}

\RequirePackage[ruled,vlined,linesnumbered]{algorithm2e}

\SetCommentSty{mycommfont}
\SetKwInOut{Input}{Input}
\SetKwInOut{Output}{Output}
\SetKw{Continue}{continue}
\DontPrintSemicolon
\SetNoFillComment
\let\oldnl\nl 
\newcommand{\nonl}{\renewcommand{\nl}{\let\nl\oldnl}} 

\AtBeginDocument{%
  \providecommand\BibTeX{{%
    \normalfont B\kern-0.5em{\scshape i\kern-0.25em b}\kern-0.8em\TeX}}}

\copyrightyear{2023} 
\acmYear{2023} 
\setcopyright{acmlicensed}\acmConference[SIGSPATIAL '23]{The 31st ACM International Conference on Advances in Geographic Information Systems}{November 13--16, 2023}{Hamburg, Germany}
\acmBooktitle{The 31st ACM International Conference on Advances in Geographic Information Systems (SIGSPATIAL '23), November 13--16, 2023, Hamburg, Germany}
\acmPrice{15.00}
\acmDOI{10.1145/3589132.3625631}
\acmISBN{979-8-4007-0168-9/23/11}

\acmConference[SIGSPATIAL '23]{31st ACM SIGSPATIAL
International Conference on Advances
in Geographic Information Systems}{November 13--16,
  2023}{Hamburg, Germany}

\begin{document}

\title{SUSTeR: Sparse Unstructured Spatio Temporal Reconstruction on Traffic Prediction}

\author{Yannick W\"olker}
\affiliation{%
  \institution{Department of Computer Science, Kiel University}
  \city{Kiel}
  \country{Germany}}
\orcid{0009-0006-6076-8996}
\email{ywoe@informatik.uni-kiel.de}

\author{Christian Beth}
\affiliation{%
  \institution{Department of Computer Science, Kiel University}
  \city{Kiel}
  \country{Germany}}
\orcid{0000-0003-3313-0752}
\email{cbe@informatik.uni-kiel.de}

\author{Matthias Renz}
\affiliation{%
  \institution{Department of Computer Science, Kiel University}
  \city{Kiel}
  \country{Germany}}
\orcid{0000-0002-2024-7700}
\email{mr@informatik.uni-kiel.de}

\author{Arne Biastoch}
\affiliation{%
  \institution{GEOMAR Helmholtz Centre for Ocean Research}
  \institution{Kiel University}
  \city{Kiel}
  \country{Germany}}
\orcid{0000-0003-3946-4390}
\email{abiastoch@geomar.de}

\begin{abstract}
Mining spatio-temporal correlation patterns for traffic prediction is a well-studied field. 
However, most approaches are based on the assumption of the availability of and accessibility to a sufficiently dense data source, which is rather the rare case in reality. 
Traffic sensors in road networks are generally highly sparse in their distribution: fleet-based traffic sensing is sparse in space but also sparse in time.
There are also other traffic application, besides road traffic, like moving objects in the marine space, where observations are sparsely and arbitrarily distributed in space.
In this paper, we tackle the problem of traffic prediction on sparse and spatially irregular and non-deterministic traffic observations.
We draw a border between imputations and this work as we consider high sparsity rates and no fixed sensor locations.
We advance correlation mining methods with a Sparse Unstructured Spatio Temporal Reconstruction (SUSTeR) framework that reconstructs traffic states from sparse non-stationary observations.
For the prediction the framework creates a hidden context traffic state which is enriched in a residual fashion with each observation.
Such an assimilated hidden traffic state can be used by existing traffic prediction methods to predict future traffic states.
We query these states with query locations from the spatial domain.
\end{abstract}




\maketitle

\section{Introduction}

\begin{figure*}
    \centering
    \includegraphics[width=\textwidth]{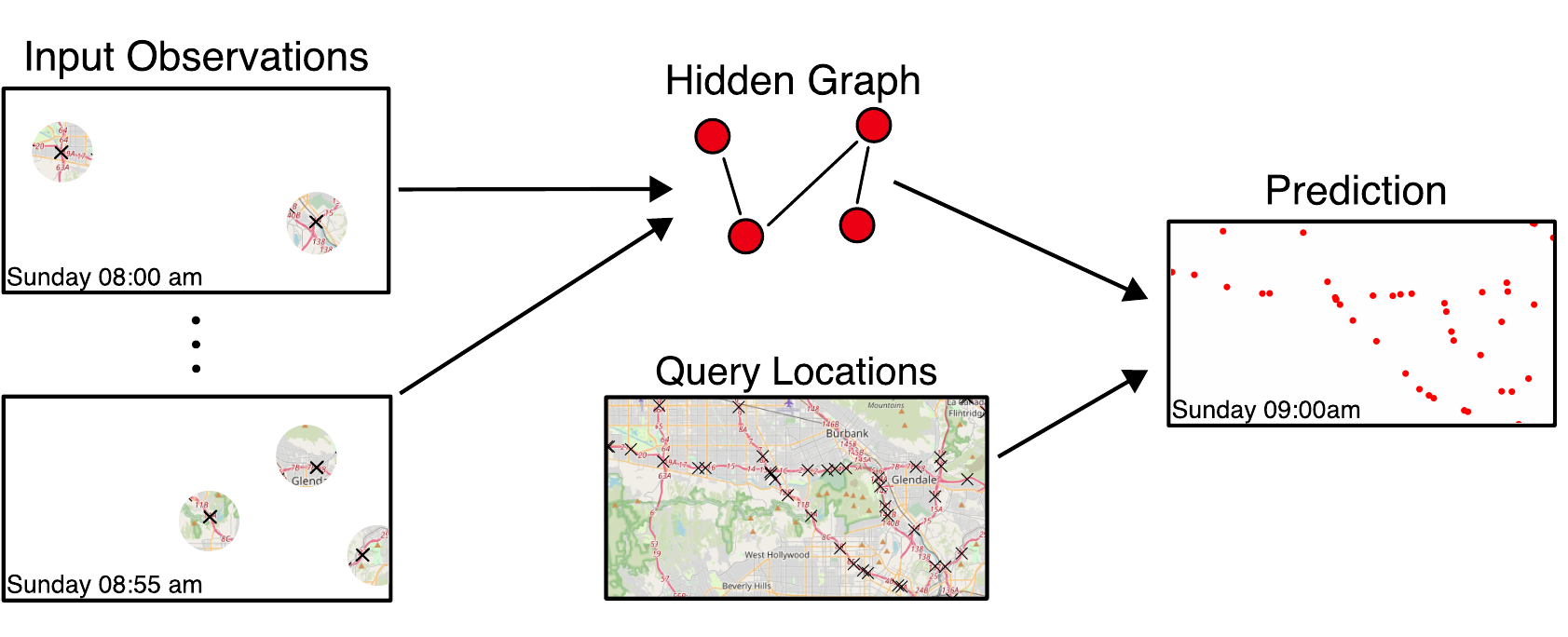}
    \caption{Showing the novel problem statement applied to traffic prediction use case. Multiple unstructured observations from the past are used to reconstruct a hidden traffic state from which a full traffic state is forecast with a set of query locations. }
    \label{fig:intro}
\end{figure*}

Forecasting the traffic in the near future is an important task for city management.
Data from the near past is used to predict future traffic states with spatio-temporal Graph Neural Networks \cite{bui22}.
Accurate prediction provides the opportunity to optimize traffic flow, reduce traffic jams and increase air quality \cite{Po19}.

While traffic prediction relies on the availability of data from traffic sensors, there exists a plethora of reasons why sensors may stop working temporarily, such as simple errors, energy saving, or overloaded communication systems.
Considering small- or medium-sized cities, the coverage of sensors may be low because the sensors are too expensive or not available.
Also, the sensors are typically static and do not adapt to changes in the traffic flow (e.g. caused by a construction site), which motivates moving sensors that for example could be mounted on cars. 
However, both missing and moving sensors introduce sparsity, since measurements may not be available for all locations at all times.
This sparsity must be explicitly addressed in traffic prediction for a realistic application scenario, which is illustrated in figure \ref{fig:intro}.
From one hour of data on Sunday morning, only few observations of the traffic state are available at each timestep.
The number of observations may differ throughout the observed time and the observation itself can be distributed arbitrarily in the city. 
We assume a relatively low number of sensors to account for resource saving and sensor failure in our proposed framework SUSTeR.
The task is to predict the dense traffic state one timestep after the observations at all possible sensor locations.
We study this problem on the traffic dataset Metr-LA and PEMS-BAY to test our assumption that only a fraction of the sensor values would be enough for good predictions.
By modifying an existing traffic dataset, we are able to compare our results from very sparse observations to the bottom line with all information available.
A successful study will provide insights in how sensors in new cities can be reduced before installing them and further mobile sensors would save more resources and are able to adapt to new traffic situations.
We argue that in order to be adaptable to other cities and changes in traffic flows, prior information like the road network should be neglected and just the sparse observations considered.
This comes with the added benefit of making our solution applicable in regions where no openly available road network is maintained or pathways change frequently (e.g. flood areas, animal observations).

The aforementioned problem is novel and more challenging than the commonly considered traffic prediction problem, since there exist very few observations in each input sample.
Current works for the traffic prediction problem do not consider any missing values. \cite{Li2021, Shao22}
A common method among state of the art approaches is the usage of Graph Neural Networks on graphs that model the sensor network \cite{bui22}.
The values of a sensor are applied to the same graph node for each timestep which prohibits any non-stationary sensors . 
With fixed sensor locations, the resulting sensor network is highly correlated with the road network.
Streets connecting two intersections with sensors should be also an interesting point for correlations in the sensor network.
However, variable observations and high temporal sparsity rates can not be modeled adequately in a static network.
We show in our experiments that the road network has only a small influence on the traffic predictions.

Besides the traffic prediction for future timesteps, some works explore the field of traffic speed imputation \cite{Cini22, Cuza22} where missing sensor values are predicted.
But the amount of missing values is assumed to be at most 80\%, which on average are still over 40 given sensors in each timestep in the Metr-LA dataset with a total of 207 sensors.
We consider up to 99.9\% missing values which are on average 2.4 observations in each timestep that are used as input.
Such high sparsity rates drastically decrease the chance that multiple values are present in one input sample from the same sensor location, which makes it challenging to recognize and learn temporal correlations for each location on its own.

High sparsity rates (>95\%) result in few sensor values, but if a reconstruction of the traffic state would be possible, we question if spatio-temporal graphs require nodes for each sensor.
In SUSTeR we utilize only a small amount of graph nodes for the encoding of information and do not relate such nodes to the sensor network.
We call this the hidden graph (see figure \ref{fig:intro}), which is still able to reconstruct the complete traffic state.
Due to the reduced number of nodes SUSTeR achieves faster runtimes, as shown in the experiments.
This hidden graph is not embedded directly in the spatial domain, which is why the assignment of observations, as well as the querying of the future traffic, is done with an encoder and a decoder, implemented as neural networks.
The decoding from the hidden graph to future values depends on a set of query locations.
Figure \ref{fig:intro} shows the query locations as given from outside and in combination with the reconstructed traffic state the future values are predicted.

To construct the hidden graph we encode observations from each timestep into from multiple graphs, one for each timestep. 
The graphs are created in a residual style and information is added to the node embeddings from the previous timesteps.
We choose this method to incorporate all timesteps equally into the hidden state because the redundant information along the past is non-existing for high sparsity rates.
From the sequence of graphs where our framework inserted the observations step by step we apply STGCN \cite{Yu18}, an algorithm for traffic prediction to find and learn the spatio-temporal correlations on our small number of graph nodes.
The first future timestep of the STGCN is our hidden graph in which the traffic state is reconstructed. 


We find in the experiments that SUSTeR outperforms the plain STGCN and modern traffic prediction frameworks like D2STGNN for high sparsity rates $(\geq 99\%)$.
This is equivalent to only $0.2$ to $2.4$ observation for each timestep on average.
SUSTeR uses fewer parameters than the baselines and can train faster and with less training data.
Our main contributions can be summarized as follows:
\begin{itemize}
    \item We introduce a sparse and unstructured variant of the traffic prediction problem with sparsity in all dimensions. The sensors report only a fraction of their values and are arbitrarily distributed in the spatial domain.
    \item We propose SUSTeR, a framework around the STGCN architecture, which maps sparse observations onto a dense hidden graph to reconstruct the complete traffic state.
    Our code is available at github.\footnote{https://github.com/ywoelker/SUSTeR}
    \item We conducts experiments that show that SUSTeR outperforms the baselines in very sparse situations ($\geq 95\%$) and has a competitive performance in low sparsity rates.
\end{itemize}

\section{Problem Definition}
\label{sec:problem}

Our problem is a variant of the multivariate time series prediction that is commonly considered for traffic forecasting. 
We formulate a more general problem that can handle high sparsity and is not restricted by spatial structures such as roads. 
This flexible formulation enables the usage of unstructured observations that are moving and are inconsistent in time. 
Important for a proper reconstruction within sparse and unstructured observations is the detection of latent dependencies in a series of observations which can be applied across the sparse samples.
There is a high chance that those dependencies will only be partially part of a single sparse training sample which is in contrast to the common traffic prediction.

Let $\mathcal{S}$ be a spatial domain, which can be continuous or discrete, and $\mathcal{T} = \{t_0, \dots, t_{m-1}, t_{m}\}$ a temporal domain with $m+1$ discrete timesteps.
Then let $\mathbf{O}$ be a set of observations in $\mathcal{S}$ throughout the first $m$ timesteps $\mathcal{T}_{obs} = \mathcal{T} \setminus \{t_{m}\}$.
Each observation is a tuple of a time $t$, a spatial position $s$, and a vector of the observed values $\bm{y} \in \mathbb{R}^{d_f}$ with dimension $d_f \in \mathbb{N}$ :
\begin{equation}
    \mathbf{O} \subseteq \mathcal{T}_{obs} \times \mathcal{S} \times \mathbb{R}^{d_f}
\end{equation}
The high flexibility of the problem is expressed by the unstructured observations, a varying amount of observations for each timestep $t$, and observation positions which can change for each $t$.
High sparsity within the spatial and temporal domain can be modeled by this problem definition because it is possible to have few observations in a region that do not have to appear in any other past or future timestep. 

The goal is to predict for a subset of query locations $Q \subseteq \mathcal{S}$ the future values $\hat{\bm{y}} \in \mathbb{R}^{d_f}$ for the timestep $t_m \in \mathcal{T}$.
Therefore we learn a function $F: \mathcal{P}(\bm{O}) \times \mathcal{S} \to \mathbb{R}^{d_f}$ which predicts the future timestep from the observations $\mathbf{O}$ and the target position $s$. 
\begin{align}
    \hat{\bm{y}} = F(\bm{O}, s) ,& &  \forall \, s \in Q    
\end{align}
Note that $Q$ can be either a set of independent points of interest or a regular grid as frequently considered in previous works.
The problem challenges possible solutions to gather the information of the observations scattered across the timesteps $\mathcal{T}_{obs}$ and combine them to a reconstructed dense traffic state.
The combination is required because the observations can be arbitrarily scattered in space and in time.
Specifically, in the case of very sparse observations, a possible solution will have to learn spatio-temporal correlations, which are only partly represented in a single sample $o$ or are split between multiple samples and then fused by training over the complete data set.
For such a problem it is most important to fuse the data from previous sparse times steps for the prediction in $t_m$. 
While other problem definitions use nearly complete traffic states at each timestep as input, this problem definition forces the algorithm to exploit the temporal structures even more when the data is sparse.

Within common traffic prediction methods the road network and the information about the fixed sensor locations are often utilized, which we count as prior knowledge because the information is not dependent on the observations.
From a broader perspective such a usage of the road network to design or guide the graph graph construction as in \cite{Yu18, Zhou20} is an example of great informed machine learning algorithms \cite{vonrueden2023}.
Furthermore, our definition targets a more general case, where observations are not restricted to roads, but distributed in space without prior knowledge of the spatial structure. 
The road network is, at least in Static Graph solutions (see section \ref{sec:related}), explicitly incorporated, because it is freely available prior information.
Due to these requirements the solutions are strongly tied to the car traffic prediction task as other traffic predictions like ships or planes are not restricted in this way.
Considering traffic prediction in smaller cities with a lot fewer sensors on the road, we could handle moving sensors, e.g. sensors installed in cars.
This introduces a new complexity to the problem definition because hidden spatial structures have to be learned additionally.

\section{Method}
\label{sec:method}

Our goal is to infer from sparse observations which are spatially distributed a complete traffic state also for spatial locations where no observations are available.
Therefore we need to learn and utilize spatio-temporal correlations that describe the state of unobserved regions from just a few present observations.
In section \ref{subsec:imagine} we describe how we achieve a latent dense spatial description for a traffic state represented by the input sample. 
The following section \ref{subsec:merge} describes how we create a modeled graph from these latent descriptions and use state-of-the-art methods to predict a future latent description. 
Finally, we query our latent description with query locations to receive values of interest, which is described in section \ref{subsec:query}.

\begin{figure*}
    \centering    \includegraphics[width=\textwidth]{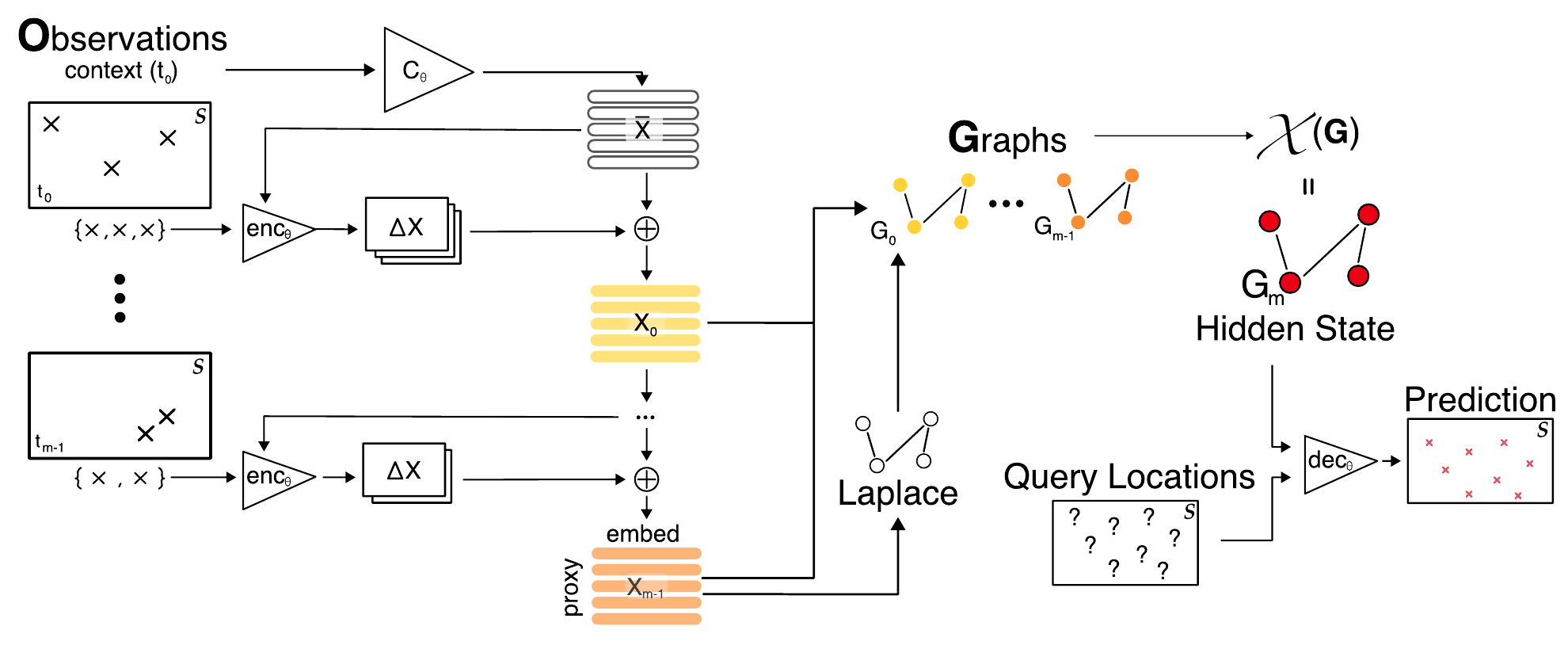}
    \caption{Framework of SUSTeR with an observation encoding, a residual architecture for hidden traffic state reconstruction with a variable amount of observations and a decoding from the dense hidden traffic state into the original space.}
    \label{fig:architecture}
\end{figure*}

\subsection{Reconstruction of Traffic State}
\label{subsec:imagine}

We want SUSTeR to learn complex spatio-temporal correlations across multiple input samples which we aim to achieve with explicit spatio-temporal correlations.
This is important because due to the sparseness of the observations those correlations are not completely represented in a single input sample.
A graph is a good representation of locations connected by spatio-temporal correlations as it was done in many works before \cite{Zhou20, Yu18, Shao22}.
In early approaches each sensor location was modeled as a single graph node \cite{Li2018} while later work grouped similar locations to districts to learn more abstract correlations \cite{Li2021}. 
We extend this idea and use a fixed number of graph nodes set $V$. 
However, these do not correspond to a specific location or sensor and the information of the observations can be assigned freely to those nodes making the usage as flexible as possible.
Such a strategy gives the framework a higher degree of freedom to find similar regions with similar correlations and assign those to the same node.
From the point of our introduced problem in section \ref{sec:problem}, the observations can have variable locations and their positions can be unique throughout the entire data, which is why we need such a flexible solution.
Each node $v \in V$ is assigned to a row in $X_{i} \in \mathbb{R}^{|V| \times d_e}, \, t_i \in \mathcal{T}_{obs}$ which contains the embedding vector with $d_e \in \mathbb{N} $ dimensions and is used to encode latent information to all spatial points connected to this graph node.
The connectivity between the introduced graph nodes is described further in section \ref{subsec:merge}.

To tackle the problem of distributed correlations in the training data we introduce a context that can be used to remember similarities across training samples.
Context information in a spatio-temporal learning setup describes the overall environment.
For example, at Sunday 08:00 am the relevant context could be the weekend.
The temporal context models a broader influence on the traffic state which can differ greatly between weekends and weekdays.
We model the context as a function $C_{\theta}: \mathcal{T} \to \mathbb{R}^{\mid V \mid \times d_e}$ that maps the temporal context information of $t_0$, to our graph nodes as a bootstrap of the node embedding for estimating the initial traffic state $\widebar{X}$.
\begin{equation}
    \widebar{X} = C_{\theta}(t_0)
    \label{eq:mean_state}
\end{equation}
We mark all functions with learnable weight with $\theta$ representing the weights.
The context $\widebar{X}$ is consecutively assimilated to the following observations $\mathbf{O}$, as illustrated in figure \ref{fig:architecture}.
We use a residual structure to enrich $\widebar{X}$ in each timestep with the observations, adding information and reducing uncertainty.
With the function $enc_{\theta}: \mathbf{O} \times \mathbb{R}^{|V| \times d_e} \to \mathbb{R}^{|V| \times d_e}$ we compute the change $\Delta X$ that a given observation $o$ imparts on the previous traffic state $X$:
\begin{equation}
    \Delta X = enc_{\theta}(o, X) = sample_{\theta}(s) \cdot inf_{\theta}(o, X)^T,
\end{equation}
where $sample_{\theta}: \mathcal{S} \to \mathbb{R}^{|V|}$ creates a one-hot assignment to select a graph node, and $inf_{\theta}: \mathbf{O} \times \mathbb{R}^{|V| \times d_e} \to \mathbb{R}^{d_e}$ contains the residual information.
Please note the observations within a single timestep can not influence each other but can utilize the past encodings.
This allows for an architecture handling various amounts of observations throughout the time in a single sample.
From the temporal node changes $\Delta X$ we create a sequence of graph node embeddings $(X_0, \dots, X_{m-1})$, which contains the residual information accumulated over all observations.
The state changes $\Delta X$ from all timesteps are aggregated as follows:
\begin{align}
X_i = 
\begin{cases}
    \widebar{X} + \,\, \mathlarger{\sum}\limits_{j=0}^{i-1} \, X_j + \mathlarger{\sum}\limits_{o \in O_i} enc_{\theta}(o, X_{i-1}) & i > 0\\[15pt]
    \widebar{X} + \mathlarger{\sum}\limits_{o \in O_i} enc_{\theta}(o, \widebar{X}) & i = 0
\end{cases}
\label{eq:Vi}
\end{align}

\subsection{Merging Graph Information}
\label{subsec:merge}

From the node embeddings $(X_{0}, \dots, X_{m-1})$ we define a sequence of graphs $\bm{G} = (G_{0}, \dots, G_{m-1})$, which have a common set of nodes $V$, and a common adjacency matrix $A$ representing weighted edges:
\begin{align}
    G_i = \left(V, A, X_i \right), & &  A \in \mathbb{R}^{|V| \times |V|}
\end{align}
In the following, we discuss how to learn $A$ from the node embeddings.
We use a self-adaptive adjacency matrix for SUSTeR, where $A$ is learned by the architecture itself because recent work showed that an adaptive adjacency matrix outperforms static road network adjacency \cite{Lan22, Wu2019, Bai20}.
Intuitively, two nodes in a spatio-temporal graph should have a strong edge when they are strongly correlated.
We adopt the idea from Bai et al. \cite{Bai20} and calculate the connectivity directly as the Laplacian from the similarity of the node embeddings.
To compute the Laplacian we use the last element of the node embedding sequence $X_{m-1}$, as it contains all the accumulated information by the nature of its construction (Eq. \ref{eq:Vi}). 
One could argue that for each graph $G_i$ the Laplacian from the node embedding $X_i$ could be used but we see the connectivity of the nodes and therefore the flow of information dependent on the overall situation which can only be represented by the finished accumulation of information.
The Laplacian $\mathcal{L}$ is computed as:
\begin{equation}
    \mathcal{L} = D^{-\frac{1}{2}} A D^{-\frac{1}{2}} = softmax\left(ReLU \left(X^{\,}_{m-1} \cdot X_{m-1}^T \right) \right)
\end{equation}

The resulting sequence of graphs $\bm{G}$ encodes all the information from the sparse observations including spatio-temporal correlations.
Note that the graphs in the final sequence only differ in their node embeddings $X_i$.

Because we transferred the sparse representation of our problem to a sequence of spatio-temporal graphs $\bm{G}$, we support any spatio-temporal graph neural network $\mathcal{X}_{\theta}$ for the aggregation of the correlation graphs, such as STGCN \cite{Yu18}, D2STGNN \cite{Shao22}, MegaCRNN \cite{jiang23}, Wavenet \cite{Wu2019}, etc.
In SUSTeR we use STGCN to provide the future graph $G_m$ at timestep $t_m$:
\begin{equation}
    G_m = \mathcal{X}_{\theta}(G_{0}, \dots, G_{m-1})
\end{equation}
The final graph $G_m = (V, A, X_m)$ encodes the reconstructed traffic state with fewer graph nodes than sensors.

\subsection{Querying of Locations}
\label{subsec:query}

A set of query locations $Q \subseteq \mathcal{S}$ can be chosen freely for traffic prediction.
The decoder function $dec_{\theta}: \mathbb{R}^{|V| \times d_e} \times \mathcal{S} \to \mathbb{R}^{d_f}$ predicts the next value $\hat{\bm{y}}$ for any location $s \in \mathcal{S}$.
\begin{align}
    \hat{\bm{y}} = dec_{\theta}(X_m, s),& &  \forall s \in Q
\end{align}
In the case of sparse traffic prediction, we choose the position of all traffic sensors from the original traffic datasets in order to evaluate the accuracy on ground truth.

\section{Evaluation}

In this section, we first introduce in section \ref{subsec:data} the dataset that we are using and describe how we modify it to fit our problem definition of Sparse Traffic Prediction. 
Section \ref{subsec:baselines} introduces the baselines and their modification for a performance comparison, which is done in section \ref{subsec:performance}.
The rest of the section is divided into a runtime comparison, a study about the necessary training set size, and in section \ref{subsec:ablation} we demonstrate the effect of important parameters.

\subsection{Data}
\label{subsec:data}

For comparison with other approaches in this field we choose the \textbf{Metr-LA} \cite{Li2018} and \textbf{PEMS-BAY} \cite{pems} dataset as a well-studied baseline for the problem of traffic prediction.
Both datasets were also used for small sparseness rates which we further elaborate on in section \ref{sec:related}.
The goal of these datasets is to predict the average speed in five-minute intervals located in two different cities Los Angeles and San Francisco respectively.
Metr-LA has data for 207 sensors from 34,272 timeslices and the PEMS-BAY 325 sensors with 52,116 samples.
As these datasets do not involve any sparseness, we need to introduce this artificially.
We select different dropout ratios and sample sensors from a uniform distribution and set their values to zero.


The choice to take a uniform sampling of all sensors was taken in absence of a more realistic strategy.
We are aware that even mobile sensors would not sample the traffic spatially uniform.
A systematic review of different skewed distributions would go beyond the scope of this work but will be pat of further investigations.
First experiments have shown that the dependency between the spatial areas with higher sampling rates and a reduction of the reconstruction error is neither linear nor symmetric.


Algorithm \ref{alg:sparsify_dataset} describes the dropout application to obtain a sparse dataset from a dense dataset with $n$ being the number of samples and $k$ the number of sensors.

\begin{algorithm}[!h]
\caption{Sparsifying Dense Traffic Dataset}
\label{alg:sparsify_dataset}
\Input{Dropout probability $P_{do} \in \left[ 0, 1 \right]$,\\dense dataset $D \in \mathbb{R}^{n \times k \times d_f}$}
    \Output{Sparse dataset $\overline{D} \in \mathbb{R}^{n \times k \times d_f}$}
    \nonl\hrulefill
    
    $m \gets \left[ 0 \right]_{n \times k \times d_f}$       \tcp*{initialize zero mask}
    \For{$i \in \lbrace 0, \dots, n-1 \rbrace$}{
        \For{$j \in \lbrace 0, \dots, s-1 \rbrace$}{
            $P_{keep} \sim U \left[0, 1\right]$ \;
            \If(\tcp*[f]{keep sensor}){$P_{keep} > P_{do}$}{
                $m\left[i, j, :\right] \gets 1$ \;
            }
        }
    }
    \Return $D \odot m$         \tcp*{return masked, sparse dataset}
\end{algorithm}

As input features we chose the original traffic volume, time of day, day of week, (cf. \cite{Li2021}) and the positional information of the sensors as longitude and latitude.
We add the positional feature because we don not use the prior knowledge of the road network which is in previous works the way of recognizing spatial distance.
This five-dimensional vector is used as input feature with a lead time of an hour (12 timesteps) and the target for the model is to predict the traffic volume for all sensors at the next timestep.

\newcommand{\res}[2]{\small{$#1 \pm #2$}}
\newcommand{\head}[1]{\multicolumn{1}{c}{#1}}

\begin{table*}[h!]
    \centering
    \begin{tabular}{cc|rrrr|rrr}
        &&\head{D2STGNN}&\head{STGCN}&\head{STGCN$_{adj}$}&\head{STGCN$^{perm}$}&\head{STGCN$^{perm}_{adj}$}&\head{D2STGNN$^{perm}_{adj}$}&\head{SUSTeR}\\
        \hline
        \multirow{3}{*}{\shortstack{10\%\\Dropout}} &MAE&\res{2.656}{0.011}&\res{2.758}{0.051}&\res{2.646}{0.016}&\res{5.195}{0.073}&\res{\textbf{4.835}}{0.071}&\res{5.136}{0.139}&\res{13.280}{0.024}\\
        &RMSE&\res{6.077}{0.056}&\res{6.192}{0.101}&\res{6.062}{0.066}&\res{11.907}{0.041}&\res{\textbf{11.429}}{0.114}&\res{11.712}{0.116}&\res{20.900}{0.068}\\
        &MAPE&\res{0.062}{0.000}&\res{0.067}{0.003}&\res{0.060}{0.001}&\res{0.117}{0.001}&\res{\textbf{0.101}}{0.002}&\res{0.110}{0.006}&\res{0.241}{0.001}\\
        \hline
        \multirow{3}{*}{\shortstack{80\%\\Dropout}} &MAE&\res{4.940}{0.124}&\res{4.214}{0.061}&\res{3.484}{0.035}&\res{5.173}{0.133}&\res{4.986}{0.046}&\res{4.993}{0.025}&\res{\textbf{3.969}}{1.985}\\
        &RMSE&\res{9.787}{0.075}&\res{9.599}{0.080}&\res{8.257}{0.038}&\res{11.910}{0.180}&\res{11.666}{0.055}&\res{11.731}{0.051}&\res{\textbf{9.287}}{4.644}\\
        &MAPE&\res{0.118}{0.002}&\res{0.118}{0.004}&\res{0.083}{0.002}&\res{0.116}{0.007}&\res{0.108}{0.002}&\res{0.109}{0.001}&\res{\textbf{0.089}}{0.044}\\
        \hline
        \multirow{3}{*}{\shortstack{90\%\\Dropout}} &MAE&\res{5.692}{0.091}&\res{4.975}{0.089}&\res{4.024}{0.022}&\res{5.288}{0.121}&\res{5.054}{0.037}&\res{5.066}{0.010}&\res{\textbf{4.963}}{0.023}\\
        &RMSE&\res{11.392}{0.152}&\res{11.084}{0.164}&\res{9.560}{0.027}&\res{12.115}{0.101}&\res{11.807}{0.032}&\res{11.827}{0.007}&\res{\textbf{11.615}}{0.041}\\
        &MAPE&\res{0.148}{0.004}&\res{0.135}{0.006}&\res{0.091}{0.001}&\res{0.122}{0.005}&\res{0.110}{0.002}&\res{0.110}{0.001}&\res{0.111}{0.001}\\
        \hline
        \multirow{3}{*}{\shortstack{99\%\\Dropout}} &MAE&\res{6.346}{1.089}&\res{6.182}{0.139}&\res{5.517}{0.085}&\res{7.379}{0.017}&\res{7.016}{3.139}&\res{5.494}{0.088}&\res{\textbf{5.274}}{0.059}\\
        &RMSE&\res{13.247}{1.042}&\res{13.386}{0.319}&\res{12.634}{0.168}&\res{14.608}{0.328}&\res{14.207}{3.343}&\res{12.508}{0.133}&\res{\textbf{12.236}}{0.107}\\
        &MAPE&\res{0.149}{0.033}&\res{0.142}{0.007}&\res{0.121}{0.003}&\res{0.189}{0.014}&\res{0.145}{0.048}&\res{0.124}{0.001}&\res{\textbf{0.120}}{0.001}\\
        \hline
        \multirow{3}{*}{\shortstack{99.9\%\\Dropout}} &MAE&\res{8.996}{0.545}&\res{11.007}{0.533}&\res{10.766}{1.451}
&\res{13.288}{0.008}&\res{13.304}{0.013} & \res{9.802}{0.227}& \res{\textbf{6.986}}{0.061}\\
        &RMSE&\res{17.397}{0.639}&\res{19.051}{0.419}&\res{18.822}{1.238}&\res{20.890}{0.026}&\res{20.943}{0.068} & \res{17.966}{0.173} & \res{\textbf{15.081}}{0.066}\\
        &MAPE&\res{0.191}{0.006}&\res{0.253}{0.004}&\res{0.239}{0.025}&\res{0.241}{0.000}&\res{0.242}{0.000} & \res{0.186}{0.004}& \res{\textbf{0.143}}{0.001}\\
        \hline
    \end{tabular}
    \caption{Performance of baselines and SUSTeR on the modified Metr-LA data set with various dropout rates. Reported are mean and standard deviation of the metrics of five runs for each cell. The two first columns show the original baselines with prior knowledge. The following columns present the independent influence of each modification to the STGCN baseline. The last three columns compare the performance of the modified baselines with SUSTeR.}
    \label{tab:baseline}
\end{table*}

\subsection{Implementation and Resources}
\label{subsec:impl}

We implemented our code in python using the PyTorch library \cite{paszke19} for the Neural Network.
For our baselines, we used the implementation in the library BasicTS at github \cite{liang23}.
Since SUSTeR leverages a spatio-temporal GNN, we modified their implementation of STGCN to use it as the inner spatio-temporal Neural Network $\mathcal{X}_{\theta}$ (cf. section \ref{subsec:merge}).
Specifically, we introduce a second input, the Laplacian matrix, besides the graph node features to replace the static matrix.

We choose the following hyperparameters which we found by a grid search: 
The Adam optimizer \cite{kingma2014adam} is executed with a learning rate of $5e^{-4}$ and a $L_2$-loss for the weights with $\beta = 1e^{-5}$.
A batch of the training data contains 32 samples and is shuffled throughout the epochs.
We divide our data into training (70\%), validation (10\%), and test (20\%).
Each experiment is trained on 50 epochs and we test the model from the epoch with the best validation metric.
Within the training and the validation, the loss and metric is the mean absolute error (MAE) and we report the performance additionally as root mean square error (RMSE) and mean absolute percentage error (MAPE).
All baselines were executed with the parameters proposed in their respective publications for the Metr-LA dataset.

All functions mentioned in section \ref{sec:method} with learnable parameters are neural networks mostly built of fully-connected layers and a ReLU \cite{nair2010rectified} activation function.
We denote the fully-connected layers with $FC^d$, with $d \in \mathbb{N}$ being the output dimension and use $\sigma$ as the ReLU activation function.
The individual functions are detailed below:
\begin{align*}
    C_{\theta} & : \,\, t_0 \to FC^{d_e} \to \sigma \to FC^{d_e \times \mid V \mid} \\
    inf_{\theta} & : \,\, concat(X,o) \to FC^{2d_e} \to \sigma \to FC^{2d_e}\to \sigma \to FC^{d_e}\\
    sample_{\theta} & : \,\, s\to FC^{\mid V \mid} \to \sigma \to FC^{\mid V \mid}\to softmax\\
    dec_{\theta} & : \,\, concat(X,s) \to FC^{256} \to \sigma \to FC^{128} \to FC^1 
\end{align*}
The graph node assignment $sample_{\theta}$ is obtained by feeding an observation position $s$ to a multi-layer perceptron (MLP).
The MLP's output is converted to a distribution with the softmax function, from which a single graph node index is sampled.
This index is one-hot encoded into a vector of length $|V|$.

All experiments were executed on a HPC cluster node with 16 allocated cores of an Intel Xeon Gold 6226R, with 16GB of RAM, and an NVIDIA Tesla V100-GPU.
We used CUDA version 11.1.0 together with cuDNN in version 8.0.4.30, and PyTorch version 1.10.

\subsection{Baselines}
\label{subsec:baselines}

We use two baselines for the common traffic prediction problem and adapt them to our introduced version of the traffic prediction problem in section \ref{sec:problem}.
First, we select STGCN \cite{Yu18} as a competitor since SUSTeR uses STGCN in part.
Second, D2STGNN \cite{Shao22} is a recent state-of-the-art solution, which works with a diffusion approach that should be able to cover at least some missing values by design.
As both competitors are built for traffic prediction and use prior knowledge we modify both for a fair comparison which we describe in the following.

Both architectures originally use the static road network as prior information for their graph, which technically violates our problem definition.
To avoid this, we use a random adjacency matrix for both architectures, where the values are drawn at random from $\mathcal{N}(0, 1)$.
From the random matrix we compute the normalized Laplacian in the case of STGCN, and the bidirectional transition flow matrix for D2STGNN.
The random matrix is initialized once the training starts and is constant throughout the epochs.
If this modification is enabled we mark the results on the baselines as STGCN$_{adj}$.

The second assumption is the non-stationary observations. Our inputs are so sparse that the chance that in less than ten observations two are at the same location is minimal.
In the baselines, the same sensor is every time on the same input node which creates a prior knowledge that our problem definition prohibits because the input should have variable sizes with variability in the spatial domain.
Permuting the input sensors will break this static assignment which we do before processing a batch.
In exchange we provide the baselines with the location of the sensors as latitude and longitude.
We mark the permutation modification of a baseline as STGCN$^{perm}$.

These changes impair the performance of the baselines but are necessary to fit our problem definition for a fair comparison.
Table \ref{tab:baseline} shows the impact of both modifications on different dropout rates to get an impression of the complexity.
The left four columns show the D2STGNN without any modifications and the STGCN with all possible combinations.
The results show that the permutation of the input sensors decreases the performance the most although the location is then part of the input.
With higher dropout rates the performances between no and both modifications moving closer together which depicts the increasing complexity of the task with high sparsity.
Surprising is the minimal decrease of performance between (STGCN$_{adj}$) with and (STGCN) without the random adjacency matrix which raises questions regarding the importance of the road network for the STGCN approach.
This is on par with the findings in previous works \cite{Lan22, Wu2019, Bai20} and further strengthens the decision in our problem definition to omit the prior knowledge of the road network.
\begin{table}[h]
    \centering
    \begin{tabular}{cc|c|c|c}
        & & STGCN & D2STGNN & SUSTeR \\
        \hline
        \multirow{3}{*}{\shortstack{10\%\\Dropout}} & MAE & \res{\textbf{2.293}}{0.011}& \res{2.359}{0.021} & \res{4.711}{0.002}\\
        & RMSE & \res{4.532}{0.019}&\res{\textbf{4.419}}{0.028} &\res{8.203}{0.004}\\
        & MAPE &\res{\textbf{0.052}}{0.000} & \res{0.055}{0.001}& \res{0.135}{0.000}\\
        \hline
        \multirow{3}{*}{\shortstack{80\%\\Dropout}} & MAE & \res{\textbf{2.340}}{0.013} & \res{2.353}{0.016}&\res{3.183}{1.013}\\
        & RMSE &\res{4.596}{0.026} & \res{\textbf{4.455}}{0.032}& \res{5.908}{1.625}\\
        & MAPE &\res{\textbf{0.054}}{0.000} & \res{0.056}{0.001}& \res{0.085}{0.036}\\
        \hline
        \multirow{3}{*}{\shortstack{90\%\\Dropout}} & MAE & \res{2.356}{0.015} & \res{\textbf{2.348}}{0.010}& \res{2.369}{0.011}\\
        & RMSE &\res{4.617}{0.015} &\res{\textbf{4.453}}{0.006} & \res{4.584}{0.013}\\
        & MAPE & \res{\textbf{0.054}}{0.000}& \res{0.055}{0.000} & \res{0.056}{0.000}\\
        \hline
        \multirow{3}{*}{\shortstack{99\%\\Dropout}} & MAE & \res{3.377}{1.091}
        & \res{2.491}{0.020}& \res{\textbf{2.457}}{0.017}\\
        & RMSE &\res{6.164}{1.652} &\res{\textbf{4.644}}{0.025} & \res{4.720}{0.011}\\
        & MAPE &\res{0.089}{0.038} & \res{0.058}{0.000}& \res{\textbf{0.057}}{0.000}\\
        \hline
        \multirow{3}{*}{\shortstack{99.9\%\\Dropout}} & MAE &\res{4.325}{0.776} & \res{3.814}{0.933}& \res{\textbf{2.572}}{0.013}\\
        & RMSE &\res{7.623}{1.155} & \res{6.460}{1.230}& \res{\textbf{4.915}}{0.012}\\
        & MAPE &\res{0.121}{0.027} &\res{0.102}{0.033} & \res{\textbf{0.060}}{0.000}\\
        \hline
    \end{tabular}
    \caption{Performance measured of SUSTeR and the competitors on the PEMS-BAY dataset with the mean-absolute-error, root-mean-squared-error, and mean-absolute-percentage-error.}
    \label{tab:pems_bay}
\end{table}
\subsection{Performance Results}
\label{subsec:performance}
\begin{figure}[ht]
    \centering
    \includegraphics[width = \linewidth]{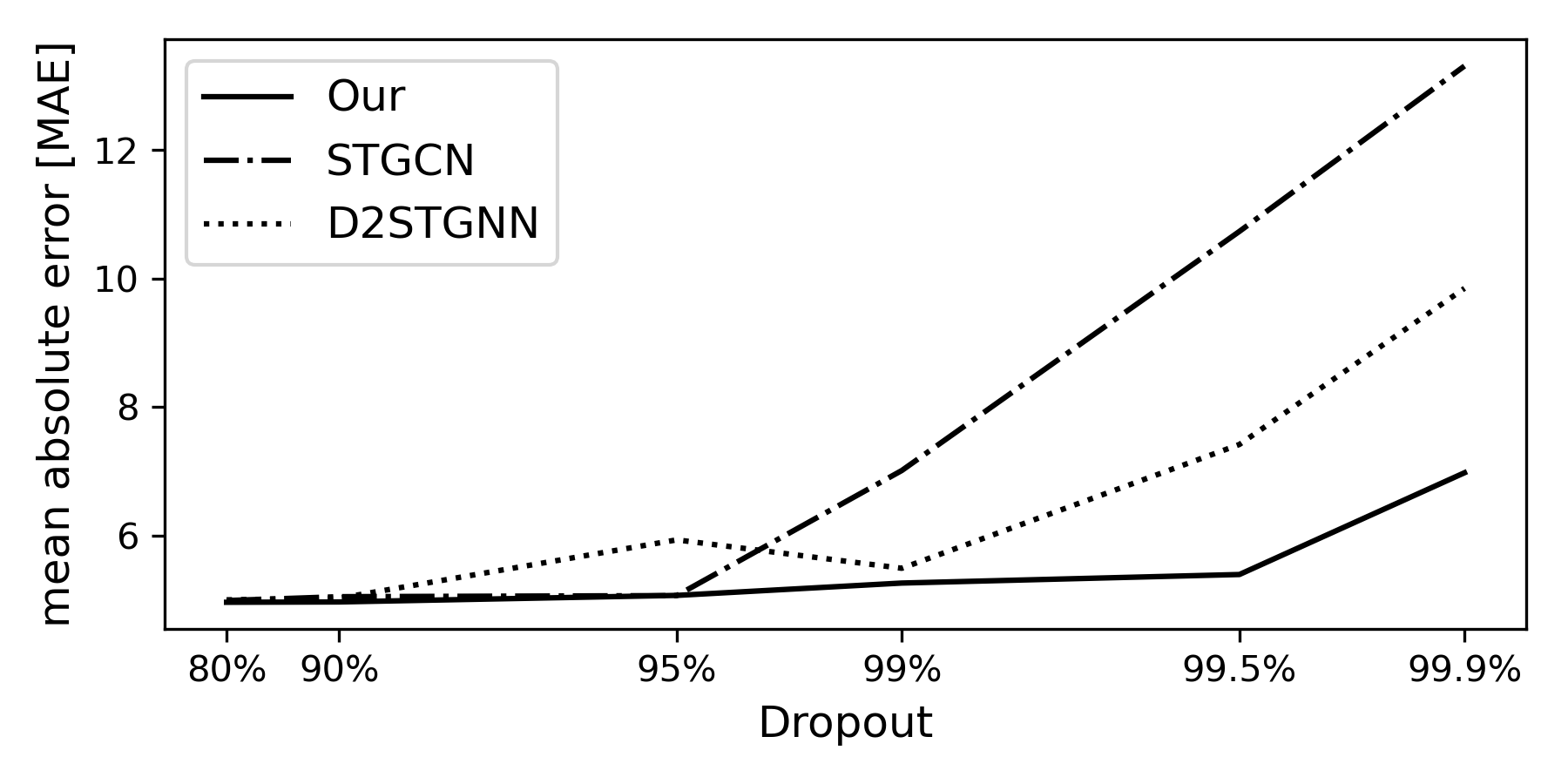}
    \caption{Mean absolute error for SUSTeR and the baselines D2STGNN and STGCN with both modifications as average over 5 runs. The dropout rates are plotted in logarithmic scaling for better visualization.}
    \label{fig:performance_mae}
\end{figure}

\begin{figure}[ht]
    \centering
    \includegraphics[width = \linewidth]{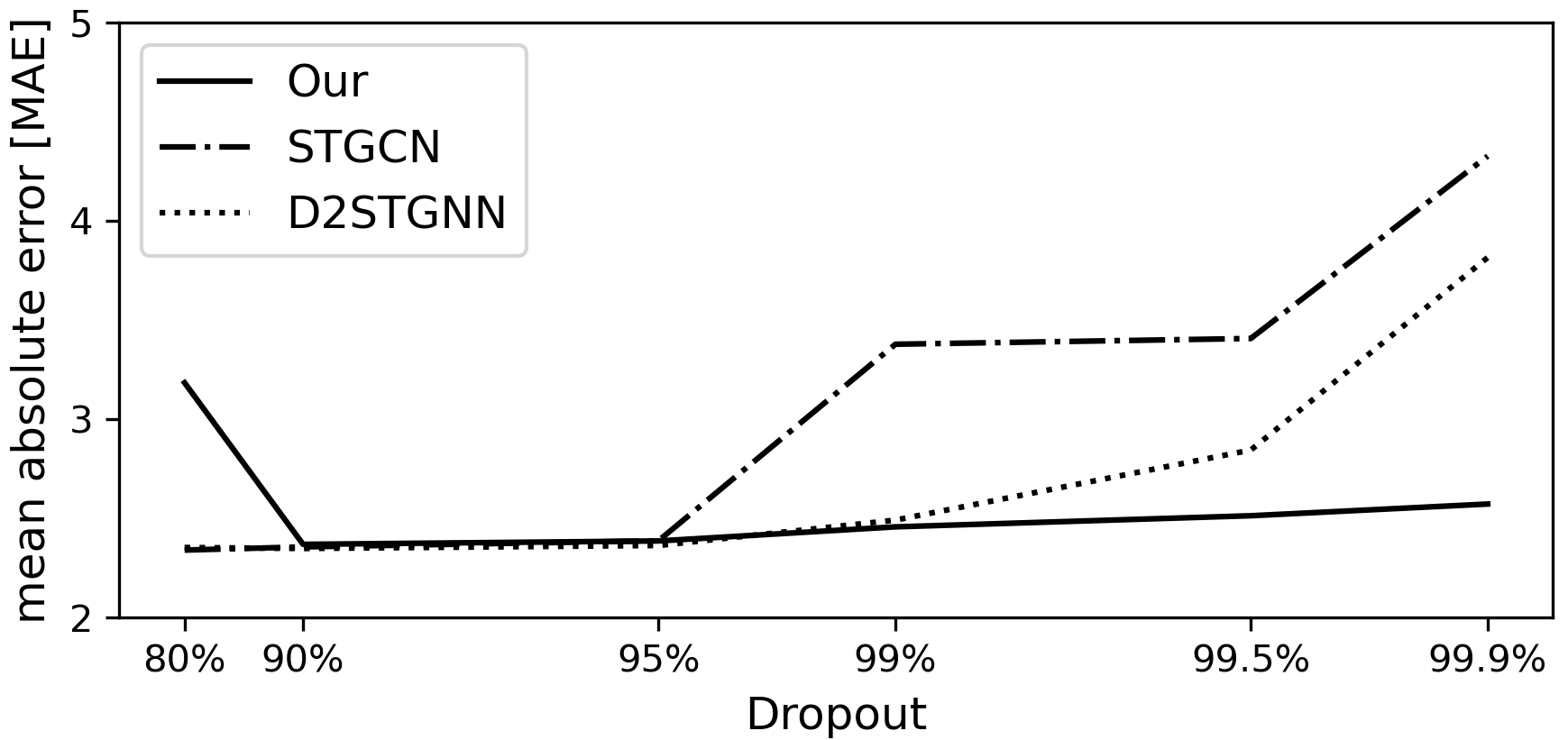}
    \caption{The mean absolute error (MAE) for the PEMS-BAY dataset evaluated on different dropout rates.}
    \label{fig:performance_mae_bay}
\end{figure}

The three right-most columns in table \ref{tab:baseline} show the performance of SUSTeR compared against both baselines with the necessary modifications (see section \ref{subsec:baselines}) for the METR-LA dataset.
The same experiment was done with the PEMS-BAY dataset and the results can be seen in table \ref{tab:pems_bay}.
We report the average metric on the test set together with the standard deviation over five runs.
While the baselines have an edge in settings with high densities, our framework clearly outperforms the baselines with very high sparsity rates of 99\% and more in all three metrics for the METR-LA dataset.
In the case of PEMS-BAY our algorithm is also superior in the very sparse regime of the experiment but stands behind for nearly complete data.
For an intuitive comparison, figure \ref{fig:performance_mae} and \ref{fig:performance_mae_bay} show the mean absolute error with additional additional dropout rates 95\% and 99.5\%.

As the main goal is to handle very sparse data we would like to refer the performance of nearly complete data to the aggregation function of $\Delta X$ and the small amount of proxies.
The length of the aggregated $\Delta X$ for a timeslice can highly vary due to the amount of observations that are present creating trouble for low dropout rates when many observations are present in the data. 
Also, in the real datasets each sensor is mapped to a unique graph node where our approach limits to only ten nodes, which is fine for high dropout rates to learn the essential spatiotemporal correlations but seems inappropriate for more information. 
This is reasonable as SUSTeR does not have the capacity to learn all correlations.

SUSTeR even has superior performance in sparse settings when the baselines are allowed access to additional knowledge in the form of the road network when comparing to single or no modified baselines.
Our framework clearly shows that for high dropouts there is more to a good prediction than the prior knowledge and that our designed architecture can handle such a high sparsity well.

\subsection{Runtime}
\label{subsec:runtime}
In this section, we compare the runtime of our approach with the chosen baselines for the METR-LA dataset.
The number of learnable parameters in all approaches is independent of the dropout rate on the input.
Therefore we picked the runtimes of the five executions for the baseline comparison in section \ref{subsec:baselines} for the 99\% dropout datasets.
Our approach uses on average 13:54 minutes ($\pm$4s) and clearly outperforms the D2STGNN architecture with 2:25 hours ($\pm$1min) for the fixed amount of 50 epochs.
Also STGCN was slower with 19:21 minutes ($\pm$6s) which is still reasonable although our framework wraps the STGCN architecture.
SUSTeR uses only half of the layer sizes of the original STGCN (see section \ref{subsec:ablation}) which reduces the number of parameters and we input only few graph nodes in comparison to all sensors in the original work.
Fewer nodes reduce the amount of convolutions that are needed within the Graph Convolution Layers (GCN) and therefore reduce the overall computation time.

\subsection{Training Data Size}
\begin{figure}[ht]
    \centering
    \includegraphics[width = \linewidth]{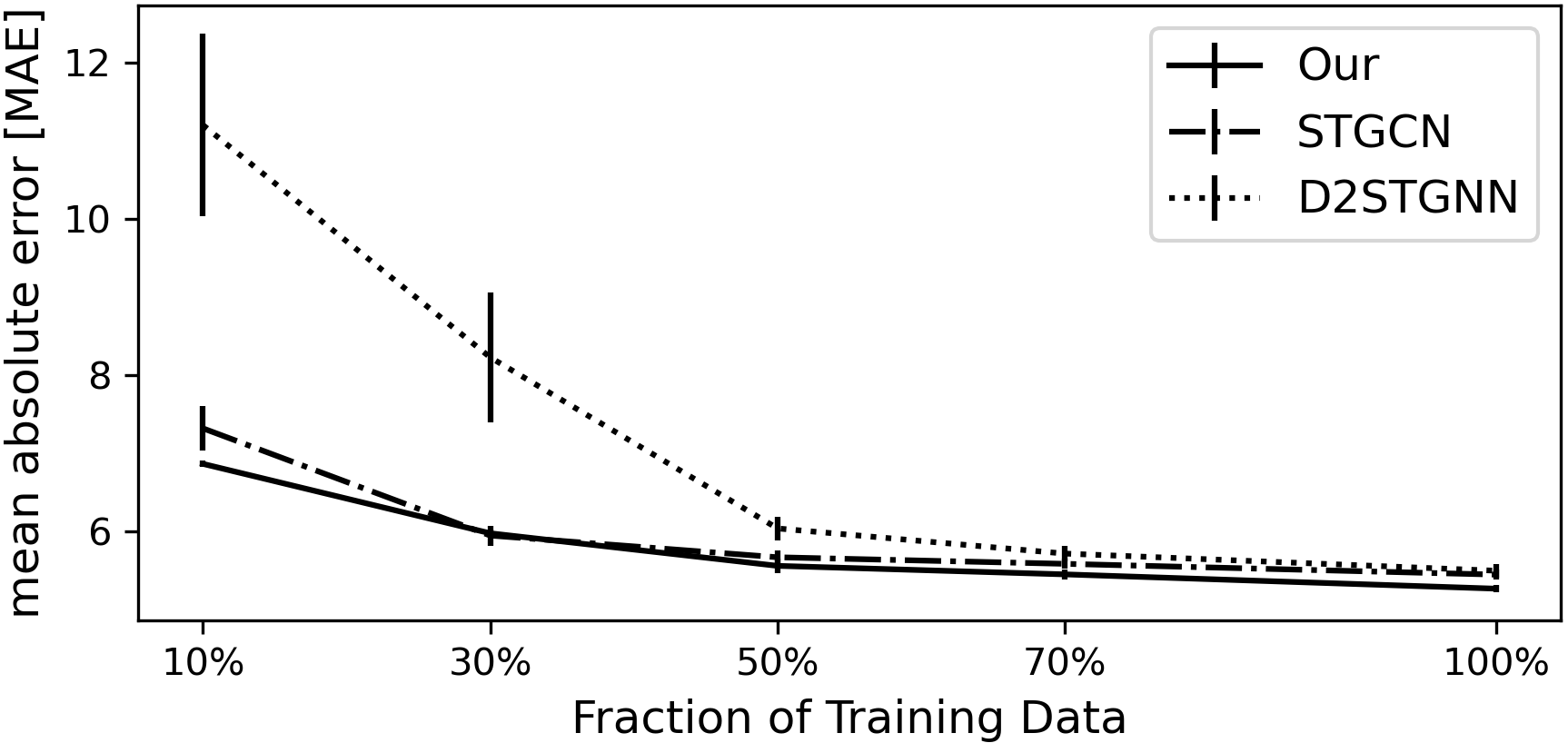}
    \caption{Training only on a fraction of the training set while keeping validation and test set the same. Executed with a 99\% dropout rate.}
    \label{fig:fraction_training_data}
\end{figure}

We evaluate the robustness of our approach with different sizes of training data to test also this dimension of sparseness along the training samples.
We keep the validation and test data fixed and take only the first 10\% to 70\% from the overall training data.
Figure \ref{fig:fraction_training_data} shows the mean absolute error of the test set with our framework and both modified baselines on the 99\% dropout dataset.
With less than 50\% of the training data we clearly outperform D2STGNN and are slightly better than the STGCN baseline.
It is also worth pointing out that at 50\% training data, SUSTeR remains competitive to the baseline that have been trained on 100\% of the training data.
Above the 50\% training data it is not possible for STGCN to draw an advantage out of more data while SUSTeR is still improving.
This shows that our framework can better learn the spatio-temporal correlations which are scattered across multiple training samples due to the high sparsity.

\subsection{Ablation Study}
\label{subsec:ablation}

To study the impact of the subparts and parameters in our framework we conduct ablation experiments on various aspects of SUSTeR.

\paragraph{\textbf{Graph nodes and Embedding}} 
A crucial parameter of the framework is the number of graph nodes $|V|$ and the embedding dimension $d_e$ which are used for the hidden traffic state.
Both parameters have a direct influence on the number of parameters in the framework and should be carefully chosen to reduce training time while still keeping a good performance.
Table \ref{tab:proxyembed} shows the results of a comparison on the 99\% dropout dataset as the mean absolute error over three runs with standard deviation.
The best values are achieved with the combinations 10 graph nodes and 32 hidden embedding dimension or 25 graph nodes and half of the embedding dimension. 
We also note that a selection of 10 to 25 graph nodes is a good decision, even a very high number of graph nodes does not increase the performance 

\begin{table}[ht]
    \centering
    \begin{tabular}{c|cccc}
        \multicolumn{1}{c}{}& \multicolumn{4}{c}{$d_e$}\\
        $|V|$ & 8 & 16 & 32 & 64 \\
        \hline
         1  & \res{5.499}{0.029}& \res{5.400}{0.078}& \res{5.317}{0.028}& \res{5.373}{0.075}\\
         5  & \res{5.329}{0.051}& \res{5.283}{0.017}& \res{5.268}{0.065}& \res{5.361}{0.045}\\
         10 & \res{5.283}{0.055}& \res{5.310}{0.090}& \res{\textbf{5.255}}{0.030}& \res{5.305}{0.015}\\
         25 & \res{5.302}{0.060}& \res{\textbf{5.244}}{0.020}& \res{5.269}{0.032}& \res{5.298}{0.041}\\
         50 & \res{7.240}{2.527}& \res{5.312}{0.035}& \res{5.306}{0.018}& \res{5.368}{0.097}\\
    \end{tabular}
    \caption{Different number of graph nodes and embedding dimensions tested on the 99\% dropout data with the mean absolute error. Each configuration was executed 5 times.}
    \label{tab:proxyembed}
\end{table}


\paragraph{\textbf{STGCN Size}}
A central component in SUSTeR is the spatio-temporal correlation mining module $\mathcal{X}$.
The used module (STGCN) is designed for a spatio-temporal graph with the same node count as the number of sensors.
In our approach we use only a fraction of the possible sensors for our hidden traffic representation.
Therefore we evaluate multiple sizes of parameter sets of STGCN, specifically full-size, a half, a quarter, and the absence of STGCN by using fractions of the hidden dimensions.
When disabling the STGCN we take the average of the node embeddings $\{X_0, \dots X_{m-1}\}$ across the $m$ past timesteps to create the $X_m$ embedding directly.
Table \ref{tab:factor} shows that the half-size network is favorable, especially when considering that this shortens the training time compared to the full-size network.
The disabling of of the STGCN shows worse results, which clearly justify the usage of such an architecture in our framework. 
Barring that version, there is little spread to be observed, which indicates robust performance.

\begin{table}[ht]
    \centering
    \begin{tabular}{r|ccc}
        $factor$& MAE&RMSE&MAPE\\
        \hline
        1.00 & \res{5.287}{0.103} & \res{12.256}{0.189} & \res{3.444}{0.092} \\
        0.50 &\res{\textbf{5.257}}{0.043} & \res{\textbf{12.254}}{0.092} & \res{\textbf{3.409}}{0.059} \\
        0.25 & \res{5.320}{0.043} & \res{12.255}{0.044} & \res{3.421}{0.051} \\
        None & \res{6.212}{0.071} & \res{13.498}{0.151} & \res{4.090}{0.121}\\
    \end{tabular}
    \caption{Experiment on the 99\% dropout data with different layer sizes factors of the inner STGCN. None is the replacement of the STGCN with an average aggregation.}
    \label{tab:factor}
\end{table}

\begin{figure}[ht]
    \centering
    \includegraphics[width= \linewidth]{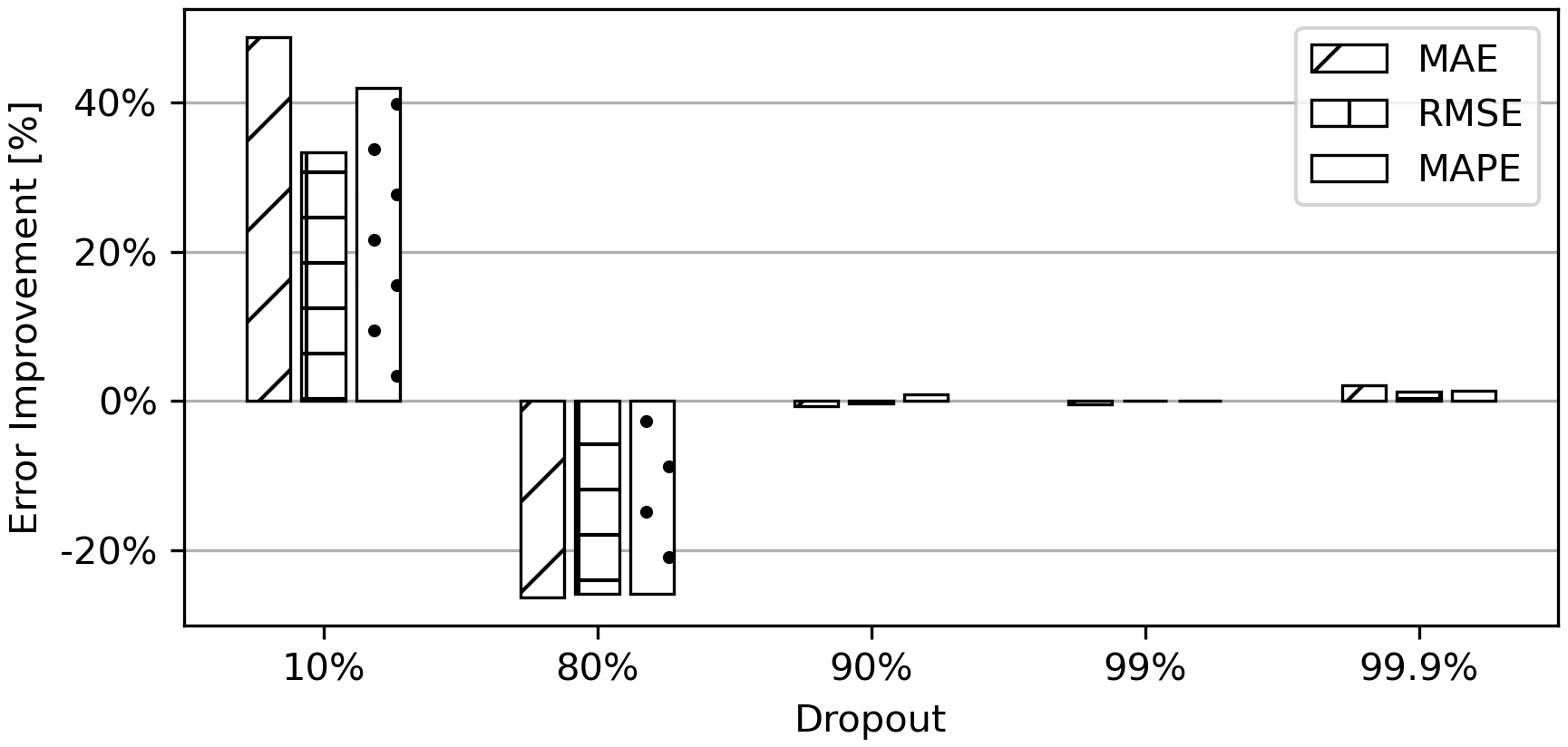}
    \caption{Relative improvement of error metrics when using D2STGNN instead of STGCN as correlation mining module $\mathcal{X}$ within SUSTeR.}
    \label{fig:improvement}
\end{figure}

\paragraph{\textbf{STGCN vs. D2STGNN}}
In this experiment we want to show what happens if we exchange the inner correlation mining module $\mathcal{X}$ from the chosen STGCN to the also tested D2STGNN.
Beforehand note the architecture of D2STGNN is in contrast to STGCN not designed for all different kinds of the spatio-temporal problems as also the \textit{time of day} and \textit{day of week} are strongly interwined into the architecture.
We changed to original implementation by replacing the static graph weights with the adjacency matrix that is learned by SUSTeR and kept the hidden dimension $d_e$ as well as the amount of proxies $\mid V \mid$ the same as in previous experiments.
Interestingly, the changes for 10\% and 80\% dropout are large (see figure \ref{fig:improvement}). 
One time a strong improvement, making the results nearly comparable to baseline STGCN/D2STGNN results (see table \ref{tab:baseline}), and on the other hand encountering a strong decrease in performance for $80\%$.
For higher dropouts the results are changing less then one percent and therefore we argue that SUSTeR itself is responsible for the great performance in the very sparse regime.
While the error is nearly not changed the runtime for SUSTeR with the D2STGNN as core is 38:59 ($\pm$57s) minutes about three times longer than with STGCN (see section \ref{subsec:runtime}).
From the minimal changing error, the longer runtime and the less modifications to the algortihm we decided to create SUSTeR around the STGCN algorithm.

\section{\textbf{Related Work}}
\label{sec:related}

The field of multivariate time series prediction is well-studied with the use-case of forecasting the traffic speed in cities given some observation history from street sensors \cite{Yu18, Li2021, Lan22}. 
State of the art works achieve satisfying results on the prediction of the next future timesteps from an equal amount of timesteps in the past.
In a timestep the values of all sensors are available and can be used to exploit spatio-temporal correlations for a prediction of all sensors in the future timesteps. 
This problem is continuously adapted to new problem variants, like Ramhormozi et al. introduced the forecasting of trucks dependent on weather conditions \cite{Ramhormozi22}.
To solve the problem of traffic prediction, spatio-temporal patterns between the traffic sensors play a crucial role.
Existing approaches can be divided into approaches that model the spatial dependencies in graphs, and in multi-dimensional tensors \cite{choy20194d}. 
Multi-dimensional with high sparsity rates tensors would be strongly zero-inflated, which results in a large computational overhead.
We present multivariate time series prediction methods and how they model the spatial correlations in graphs to provide a background for our design choices:

\paragraph*{\textbf{Static Graphs}}
For traffic prediction benchmark sets like Metr-LA the road network is available providing the connectivity of the sensors. 
The streets and intersections are assumed as prior knowledge and are incorporated as a static adjacency matrix into the correlation mining \cite{Yu18, Zhou20}.
Such knowledge remains unchanged over the entire prediction process and is used to build the adjacency matrix of the sensor graph.
Within the graph convolutions, the information of the sensors spreads out along the road network.
The motivation is that cars and traffic events (e.g. congestion) propagate along roads which is therefore the base of information flow. 
In addition to the road network, Shao et al. \cite{Shao22} have incorporated an attention mechanism that exploits the road network to strengthen connections that are highly correlated and weaken streets that are rarely used. 
We count such methods to static graphs because they are still relying on the known street network and assume that the graph does not change.

\paragraph*{\textbf{Dynamic graphs}}
\label{subsec:dynamic}
Street networks can not capture all spatio-temporal correlations because some correlations are not dependent solely on the streets (e.g. rush hours in office districts) \cite{Li2021, Shao22}.
Also long-range correlations between distant locations can not be exploited because the information between two sensors with many hops in between will be averaged out.
Different from the static graphs, a self-adaptive transition matrix is created completely data-driven. 
One of the earliest works in the traffic forecasting context with such a dynamic graph is Graph Wavenet \cite{Wu2019}, where the similarity of node embeddings is transformed into an adjacency matrix.
Lan et al. created an architecture, which infers the important (i.e., highly correlated) sensor connections only from the sensor time series \cite{Lan22}.
This motivates our usage of a similar approach to not rely on the road network for the connection of the nodes in our hidden graph. 
A static linking between the sensors and the graph nodes can not capture observation from moving sensors. 
Li et al. advocate for a dynamic graph arguing that distant districts of a city could still be correlated, despite not being directly connected in a road network \cite{Li2021}. 
For example, two office districts are likely to have similar rush hour behavior, even if they are located in different parts of the city.
Such latent connections are not explicitly available prior to training (even when considering the road network) and need to be learned from the observations in a data-driven approach.

In dynamic graphs the connection between similar districts can be strengthened by the learning process, but the graph nodes across different districts will behave similarly because the Graph Convolutions share the same weights for all nodes, although the districts can be different.
Bai et al. approach this by creating a shared weights dictionary, which is queried by the node embedding. \cite{Bai20}
The node embedding acts as clustering, connecting similar districts through the learned adjacency matrix and applying different weights to different districts. 

\paragraph*{\textbf{Sparse Time Series Forecasting}}
The previously discussed approaches assume that the available traffic data is complete.
However, missing data due to sensor failure or lack of sensors calls for the consideration of missing values in traffic prediction.
Recent imputation strategies on sparse traffic prediction handle up to 80\% missing values \cite{Cuza22}, which can be also achieved to a certain extent by traffic prediction methods which are not designed for missing values.
Considering the resource efficiency for traffic monitoring as few sensors as possible should be deployed which raises the question if less than 20\% of the values is also suffcient for a traffic prediction.
Cuza et al. \cite{Cuza22} proposed an imputation method that learns the traffic probability distribution in four velocity bins, which is then used to sample for the missing values.
They introduced a context-aware graph convolutional network (GCN), which is able to differentiate between observed and unobserved nodes in the graph, and only considers the observed nodes as context.
Over multiple iterations, the information is passed through the unobserved nodes, whose context is updated to remember which information was used as input.
Their approach assumes a previously known, static `edge graph`, which is extracted from the road network and has the same disadvantages as the static graphs approach.
Cui et al.'s work \cite{Cui20} has a lower imputation rate of up to 40\% and solves the problem with a spectral graph Markov network which uses a Markov process to model the temporal dependencies.
Another recent imputation work `Filling the G\_AP\_S` \cite{Cini22} proposed a recurrent graph neural network approach, which uses a spatio-temporal encoder to embed the nodes into a latent space.
Their imputation is primarily based on this latent space to impute temporal correlations at a single point in space, because they argue that sensor failures in such a network are rare and will involve only single sensors.


\section{Conclusion}

This paper introduces a variant of the multivariate time series traffic prediction problem with a focus on highly sparse and unstructured observations.
To address this problem we propose SUSTeR, a framework which handles sparse unstructured observations by creating hidden graphs in a residual fashion, which are then used with a conventional spatio-temporal GNN.
SUSTeR achieves better predictions for high sparsity (80\% - 99.9\% missing data) than existing baselines and remains competitive in denser settings or even when using only half the amount of the training data.
In addition, its training is considerably faster than the next-best competitor due to a smaller model size.


\section{Future Work}

We plan to explore the interpretability within SUSTeR to obtain an intuitive understanding of the graph nodes within the hidden graph.
Small design choices are made within SUSTeR to make this possible, from observations that are not relying on each other in the same timestep, variable amounts of observations, a learnable assignment function from the observation to the hidden node, and an explicit learned laplacian matrix. 
The problem of sparse unstructured observations, which should be reconstructed into a hidden state, is present in many other domains.
In particular ocean data is a very promising application field for SUSTeR where sparse ARGO\footnote{https://argo.ucsd.edu} observations would perfectly match the problem definition to predict ocean states. 
There, observations are typically spatially and temporally sparse - comparable to the highest dropout rate in this paper - and observations are non-stationary and change their position freely.
We see SUSTeR as a bridge of the well-studied spatio-temporal mining methods into a new area of domains, in which such methods previously were not applicable.

\begin{acks}
We acknowledge support by the Helmholtz School for Marine Data Science (MarDATA) funded by the Helmholtz Association (Grant HIDSS-0005).
We acknowledge the support through the Deutsche Forschungsgemeinschaft (DFG) - Project 491008639. 
This work is supported by the KMS Kiel Marine Science – Centre for Interdisciplinary Marine Science at Kiel University.
\end{acks}

\bibliographystyle{ACM-Reference-Format}
\bibliography{sample-base}


\end{document}